\documentclass[12pt, a4paper]{article}

\usepackage[utf8]{inputenc}
\usepackage[T1]{fontenc}
\usepackage{csquotes} 

\usepackage{geometry}
\usepackage{graphicx}
\usepackage{hyperref} 
\usepackage{caption}
\usepackage{amsmath}
\usepackage{algorithm}
\usepackage{algpseudocode}
\usepackage{booktabs}
\usepackage{tabularx}
\usepackage{colortbl}
\usepackage{xcolor}
\usepackage{listings}

\definecolor{lightgray}{gray}{0.9}

\lstdefinestyle{mystyle}{
    commentstyle=\color{gray}, keywordstyle=\color{blue}, stringstyle=\color{red},
    basicstyle=\footnotesize\ttfamily, breaklines=true, frame=single,
    captionpos=b, keepspaces=true, rulecolor=\color{black}, tabsize=2
}
\usepackage[style=apa, citestyle=authoryear]{biblatex}
\let\cite\parencite 

\usepackage{authblk}

\title{Centralized vs. Federated Learning for Educational Data Mining: A Comparative Study on Student Performance Prediction with SAEB Microdata}

\author[1]{Rodrigo Tertulino}
\affil[1]{Federal Institute of Education, Science, and Technology of Rio Grande do Norte (IFRN), Software Engineering and Automation Research Laboratory - LaPEA, 59628-330, Mossoró-RN, Brazil. \authorcr \href{mailto:rodrigotertulino@ifrn.edu.br}{rodrigotertulino@ifrn.edu.br}, ORCID: \href{https://orcid.org/0000-0002-7594-9312}{0000-0002-7594-9312}}

\date{\today}

\begin{document}
\maketitle


\begin{abstract}
The application of data mining and artificial intelligence in education offers unprecedented potential for personalizing learning and early identification of at-risk students. However, the practical use of these techniques faces a significant barrier in privacy legislation, such as Brazil's General Data Protection Law (LGPD), which restricts the centralization of sensitive student data. To resolve this challenge, privacy-preserving computational approaches are required. The present study evaluates the feasibility and effectiveness of Federated Learning, specifically the FedProx algorithm, to predict student performance using microdata from the Brazilian Basic Education Assessment System (SAEB). A Deep Neural Network (DNN) model was trained in a federated manner, simulating a scenario with 50 schools, and its performance was rigorously benchmarked against a centralized eXtreme Gradient Boosting (XGBoost) model. The analysis, conducted on a universe of over two million student records, revealed that the centralized model achieved an accuracy of 63.96\%. Remarkably, the federated model reached a peak accuracy of 61.23\%, demonstrating a marginal performance loss in exchange for a robust privacy guarantee. The results indicate that Federated Learning is a viable and effective solution for building collaborative predictive models in the Brazilian educational context, in alignment with the requirements of the LGPD.
\end{abstract}


\pagebreak


\section{Introduction}
The paradigm of data-driven decision-making has permeated numerous sectors, with education emerging as a particularly promising domain for its application. The systematic analysis of large-scale educational datasets, a field often referred to as Learning Analytics, offers unprecedented opportunities to understand and improve student learning processes~\cite{Romsaiyud}. Methodologies rooted in artificial intelligence and data mining can be employed to create personalized learning pathways, develop early-warning systems for students at risk of academic failure, and provide valuable feedback to educators and policymakers~\cite{silva_sbie}. Harnessing such computational tools allows for a shift from traditional one-size-fits-all educational models to more adaptive and student-centered approaches.

Nevertheless, the vast potential of educational data mining is met with a significant and non-negotiable impediment: the imperative of data privacy. Student data is inherently sensitive, containing personal, academic, and socioeconomic information that requires stringent protection. In Brazil, the Lei Geral de Proteção de Dados (LGPD), Law No. 13,709/2018, establishes a rigorous legal framework for the handling of such information, imposing strict limitations on the collection, processing, and centralization of personal data \cite{Nascimento2023-cv}. Consequently, traditional machine learning approaches that require aggregating raw data from multiple institutions into a single server become legally and ethically unfeasible, creating a critical barrier to collaborative research and developing robust, generalizable predictive models.

To reconcile the benefits of large-scale data analysis with these stringent privacy requirements, novel computational paradigms are necessary. Among these, Federated Learning (FL) has emerged as a compelling solution \cite{DBLP:journals/corr/McMahanMRA16}. The methodology enables collaborative model training across multiple decentralized clients, in this context, educational institutions, without exchanging or centralizing the raw data~\cite{10.1145/3594300.3594312}. Instead, a global model is trained iteratively: the server sends the model to each client, each client trains the model on its local private data, and only the resulting model updates (anonymous numerical parameters) are sent back to the server for aggregation. Such a process allows for creating a robust global model that learns from the collective data of all institutions. At the same time, the sensitive student records remain securely within their original institutional boundaries.

While the application of FL in sensitive domains like healthcare is advancing, its exploration in the educational sphere remains nascent~\cite{Tao2025-db}. A significant challenge in educational FL is the statistical heterogeneity of data across different institutions, where student populations and data distributions can vary widely, an issue addressed by specialized algorithms such as FedProx \cite{Sahu2018FederatedOI}. Therefore, A considerable gap in the literature concerning the practical application and rigorous evaluation of robust FL algorithms on large-scale, real-world national educational datasets. The Brazilian Basic Education Assessment System (SAEB) presents a unique opportunity for such an investigation, offering a rich, nationwide dataset containing academic performance metrics and detailed socioeconomic information~\cite{INEP2025SAEB}.

The present study addresses this gap by conducting a comprehensive comparative analysis. Its primary objective is to evaluate the performance of a Deep Neural Network (DNN) model trained under a privacy-preserving federated scheme using the FedProx algorithm~\cite{10.1145/3679013}. To establish a robust benchmark, the performance of the federated model is compared against a powerful, centralized eXtreme Gradient Boosting (XGBoost) model trained on the entirety of the aggregated data~\cite{10.1145/3728314}. 
The present study addresses this gap by conducting a comprehensive comparative analysis designed to answer the following research questions:

\begin{enumerate}    
\item \textbf{RQ1:} To what extent can socioeconomic features within the SAEB dataset predict student academic performance using a centralized, state-of-the-art machine learning model?
\item \textbf{RQ2:} What is the predictive performance of a Deep Neural Network trained via a FL approach (FedProx) on the same prediction task, without centralizing sensitive student data?
\item \textbf{RQ3:} How significant is the performance trade-off between the privacy-preserving federated model and the centralized benchmark, and is such a trade-off acceptable for practical application in the Brazilian educational context?
\end{enumerate}

Experiments were conducted on a dataset of over two million student records from the SAEB to investigate these questions. A powerful, centralized XGBoost model was benchmarked against a DNN model trained under a privacy-preserving federated scheme. The results indicate that the federated approach achieves a predictive accuracy remarkably close to the centralized baseline, demonstrating its viability as a powerful tool for building collaborative, effective, and privacy-compliant predictive models.

\section{Reletad Works}

To contextualize the present study, a systematic search of the Scopus database, one of the largest curated abstract and citation databases of peer-reviewed literature, was conducted. The search query focused on articles containing the keywords "Learning Analytics", "Federated Learning", and "Machine Learning" to identify recent advancements in predictive modeling within education. To ensure the relevance and novelty of the review, the search was restricted to journal articles published in English within the last five years.

The literature robustly demonstrates the efficacy of ML for predicting student dropout and academic performance. Studies frequently conduct comparative analyses of various algorithms on institutional data, with ensemble and gradient boosting methods often emerging as top performers. For instance, \textcite{Delena2025} evaluated ten models on sociodemographic and academic data, identifying XGBoost as the most accurate. Similarly, using large-scale Brazilian educational data from INEP, \textcite{Teodoro2020} found that a Random Forest model achieved high accuracy in dropout prediction. Research by \textcite{Balcioglu2023} on the OULA dataset confirmed the superiority of ensemble methods. Further emphasizing boosting algorithms, \textcite{Marcolino2025} applied CatBoost to Moodle logs, enhanced by sophisticated hyperparameter optimization and data balancing techniques. A common thread in these valuable contributions is their reliance on a centralized data paradigm, where all student records must be aggregated for analysis.

Beyond model comparison, a significant research thrust involves leveraging the rich, granular data from Virtual Learning Environments (VLEs) and Learning Management Systems (LMS). The importance of novel feature engineering from clickstream and interaction data is highlighted by \textcite{Junejo2025} to enable accurate, multi-category predictions even at the early stages of a course. On the other hand, other works focus on moving from prediction to intervention. For instance, \textcite{Qadir2025} proposes an adaptive feedback system based on instance-level explorations to provide specific, meaningful feedback to learners. Similarly, \textcite{Andrade2025} integrates EDM with active learning methodologies within a recommendation system, demonstrating a proactive approach to mitigate dropout risk. These studies underscore the potential of detailed VLE data, but they are typically implemented within a single system and require centralized data access.

As predictive models become more powerful, their application in high-stakes environments necessitates greater transparency and strategic alignment. The challenge of model interpretability is addressed through Explainable AI (XAI), with techniques like SHAP being used to make predictions understandable to educators, as demonstrated by \textcite{Mastour2025} in the context of medical education. On a strategic level, integrating AI and Big Data is a key objective for national education systems. However, its implementation is fraught with challenges, including significant concerns regarding data privacy \textcite{Khan2025}. Despite the advancements in predictive accuracy and a growing awareness of ethical considerations, a critical gap persists. While the work by \textcite{Mastour2025} involves multi-institutional data, it still requires centralization. Consequently, the challenge of building collaborative, large-scale predictive models without violating data privacy regulations remains largely unaddressed in the reviewed literature, a gap our study directly confronts.

\begin{table}[tbp] 
\centering
\caption{Comparative Analysis of Related Work and Our Study's Contribution}
\label{tab:related_work}
\footnotesize 
\resizebox{\textwidth}{!}{%
\begin{tabular}{@{}lllll@{}}
\toprule
\rowcolor{lightgray} 
\textbf{Reference} & \textbf{Objective} & \textbf{Methodology} & \textbf{Dataset} & \textbf{Limitation / Research Gap} \\ \midrule
\textcite{Junejo2025} & Multi-category student performance forecasting at early stages. & 1D-CNN, ANN-LSTM, RF & OULA Dataset & Centralized data; Does not address multi-institutional privacy. \\ \\
\textcite{Delena2025} & Predict student retention using sociodemographic data. & Comparative study of 10 ML models (XGBoost). & Single University Data & Localized context; Requires centralized sensitive data. \\ \\
\textcite{Teodoro2020} & Predict dropout risk in Brazilian public higher education. & Comparative study of 5 ML models (RF). & INEP Microdata & Large-scale but centralized analysis; Does not address LGPD. \\ \\
\textcite{Marcolino2025} & Predict student dropout using Moodle log data. & CatBoost with NSGA-II optimization. & Local Moodle Logs & Limited dataset size; Centralized data assumption. \\ \\
\textcite{Balcioglu2023} & Early prediction of student performance. & Ensemble Model, NN, SVM, DT. & OULA Dataset & Single-institution dataset; Centralized data approach. \\ \\
\textcite{Mastour2025} & Predict medical students' performance with explainable AI (XAI). & Stacking Ensemble + SHAP. & Multi-university medical data. & Addresses explainability but requires data centralization. \\ \\
\textcite{Qadir2025} & Develop an adaptive feedback system for learners in an LMS. & Stacking, Capsule Network, SVM. & Simulated and Kaggle Datasets. & Focus on feedback, not privacy; Data is centralized. \\ \\
\textcite{Andrade2025} & Prevent dropout via a recommendation system. & EDM (RF) + Active Methodologies. & Local Course Data & Focus on intervention, not privacy; Small-scale, centralized. \\ \\
\textcite{Khan2025} & Analyze AI adoption for sustainable university development. & Mixed-methods; RF for prediction. & Saudi University Data & Focus on policy; Acknowledges privacy as a key risk. \\ \\ \midrule
\textbf{Our Study} & \begin{tabular}[c]{@{}l@{}}Predict student performance on SAEB Dataset \\ while ensuring data privacy.\end{tabular} & \begin{tabular}[c]{@{}l@{}}Federated Learning (FedProx) \\ vs. Centralized (XGBoost)\end{tabular} & \begin{tabular}[c]{@{}l@{}}SAEB Microdata \\ (>2M records)\end{tabular} & \textbf{\begin{tabular}[c]{@{}l@{}}Fills the gap by providing the first empirical \\ evaluation of a robust FL algorithm on \\ large-scale Brazilian educational data, \\ quantifying the performance-privacy trade-off.\end{tabular}} \\ \bottomrule
\end{tabular}%
}
\end{table}

The reviewed literature confirms a robust and mature landscape for applying ML to predict student outcomes, with ensemble and gradient boosting methods consistently demonstrating strong performance on institutional and datasets. Besides that, these studies successfully leverage detailed academic, demographic, and engagement data to build effective models. However, its fundamental reliance on centralized data architectures is a unifying limitation across this body of work. While effective for single-institution analysis, such a paradigm is increasingly untenable for the kind of multi-institutional collaboration needed to build truly generalizable models, primarily due to stringent data privacy regulations. While the strategic importance of AI is recognized and privacy is acknowledged as a barrier, a clear gap remains in the empirical application of privacy-preserving solutions. Therefore, the present study's objectives are designed to address this critical gap directly. First, our study aims to establish a high-performance baseline using a state-of-the-art centralized model on the large-scale SAEB dataset, aligning with established methodologies. Second, we seek to implement and train a robust, privacy-preserving model using FL, simulating a real-world multi-institutional scenario. Finally, our third objective is to provide a rigorous, quantitative comparison between these two approaches, offering the first empirical analysis of the performance-privacy trade-off for predictive analytics in the Brazilian educational context.

\section{Theoretical Foundations}

This section delineates the theoretical foundations underpinning the ML paradigms evaluated in this study. We first describe the traditional centralized approach, exemplified by the XGBoost algorithm, and subsequently introduce the principles of FL, with a specific focus on the FedProx algorithm designed to handle data heterogeneity.

\subsection{Centralized ML}

\begin{figure}[h!]
    \centering
    \includegraphics[width=0.5\textwidth]{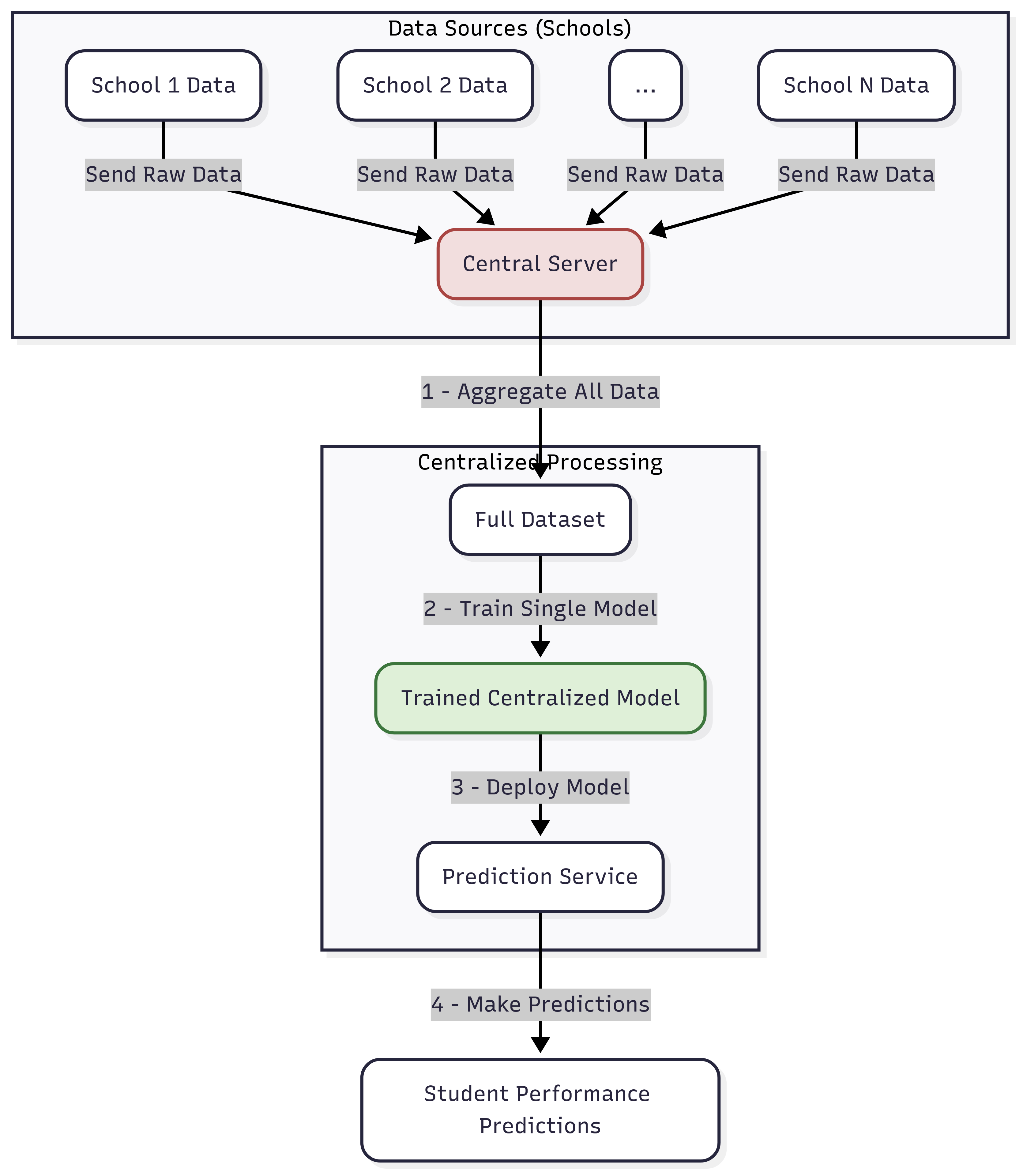}
    \caption{Centralized ML Workflow} 
    \label{fig:centralized}
\end{figure}

The conventional paradigm for training machine learning models is centralized learning. In this approach, data from all distributed sources are collected, aggregated, and stored in a single, central repository. A model is then trained on this comprehensive dataset, granting it a global view of the data distribution, a process illustrated in Figure \ref{fig:centralized}. While this method can yield high-performance models, its prerequisite of data aggregation poses significant challenges in privacy-sensitive domains like education. Transferring and storing raw student data introduce considerable risks and are often incompatible with data protection regulations such as the LGPD~\cite{SAKAMOTO20213049}.

The procedural workflow of the centralized learning paradigm used in this study is formally described in Algorithm \ref{alg:centralized}. The process begins by collecting all data from individual sources into a single, aggregated dataset, which is then used to train one monolithic model.

\begin{algorithm}[h!]
\caption{Centralized Learning Workflow}
\label{alg:centralized}
\begin{lstlisting}[language=Python]
# Input: A set of all clients (schools) K, with local datasets D_k.
# Output: A single, trained model M_final.

# --- Server Execution ---

# 1. Collect and aggregate data from all clients
D_central = aggregate_data_from(clients=K)

# 2. Preprocess the aggregated dataset
D_processed = preprocess_data(D_central)

# 3. Initialize and train the model
model = initialize_model(type="XGBoost")
trained_model = train(model, D_processed)

return trained_model
\end{lstlisting}
\end{algorithm}

\subsubsection{eXtreme Gradient Boosting (XGBoost)}

As a powerful and widely adopted algorithm for tasks involving structured or tabular data, XGBoost is an ideal baseline for centralized learning in this study~\cite{10.1145/3686852.3686881}. XGBoost is an ensemble learning technique based on the principle of gradient boosting on decision trees. It constructs a predictive model as an ensemble of weak learners, typically decision trees. The algorithm builds these trees sequentially, where each new tree is trained to correct the errors made by the ensemble of previously trained trees.

The core of XGBoost's effectiveness lies in its objective function, which is optimized at each step of the tree-building process. The objective function balances model accuracy with complexity, thereby preventing overfitting. At a given step $t$, the objective function is defined as:

\begin{equation}
\label{eq:objective_function}
Obj^{(t)} = \sum_{i=1}^{n} l(y_i, \hat{y}_i^{(t)}) + \sum_{k=1}^{t} \Omega(f_k)
\end{equation}

Where $l(y_i, \hat{y}_i^{(t)})$ is the loss function that measures the discrepancy between the true label $y_i$ and the prediction $\hat{y}_i^{(t)}$ for the $i$-th instance. The term $\sum \Omega(f_k)$ is a regularization component that penalizes the complexity of the models. For decision trees, the complexity is defined as:

\begin{equation}
\label{eq:regularization}
\Omega(f) = \gamma T + \frac{1}{2}\lambda ||\omega||^2
\end{equation}

Here, $T$ is the number of leaves in the tree, $\omega$ represents the vector of scores on the leaves, and $\gamma$ and $\lambda$ are regularization parameters that control the penalty for the number of leaves and the magnitude of the leaf weights, respectively. By minimizing this objective function, XGBoost produces a highly accurate and well-generalized model, representing a state-of-the-art benchmark for centralized performance~\cite{10.1145/3578339.3578352}.

\subsection{Federated Learning}

\begin{figure}[h!]
    \centering
    \includegraphics[width=0.5\textwidth]{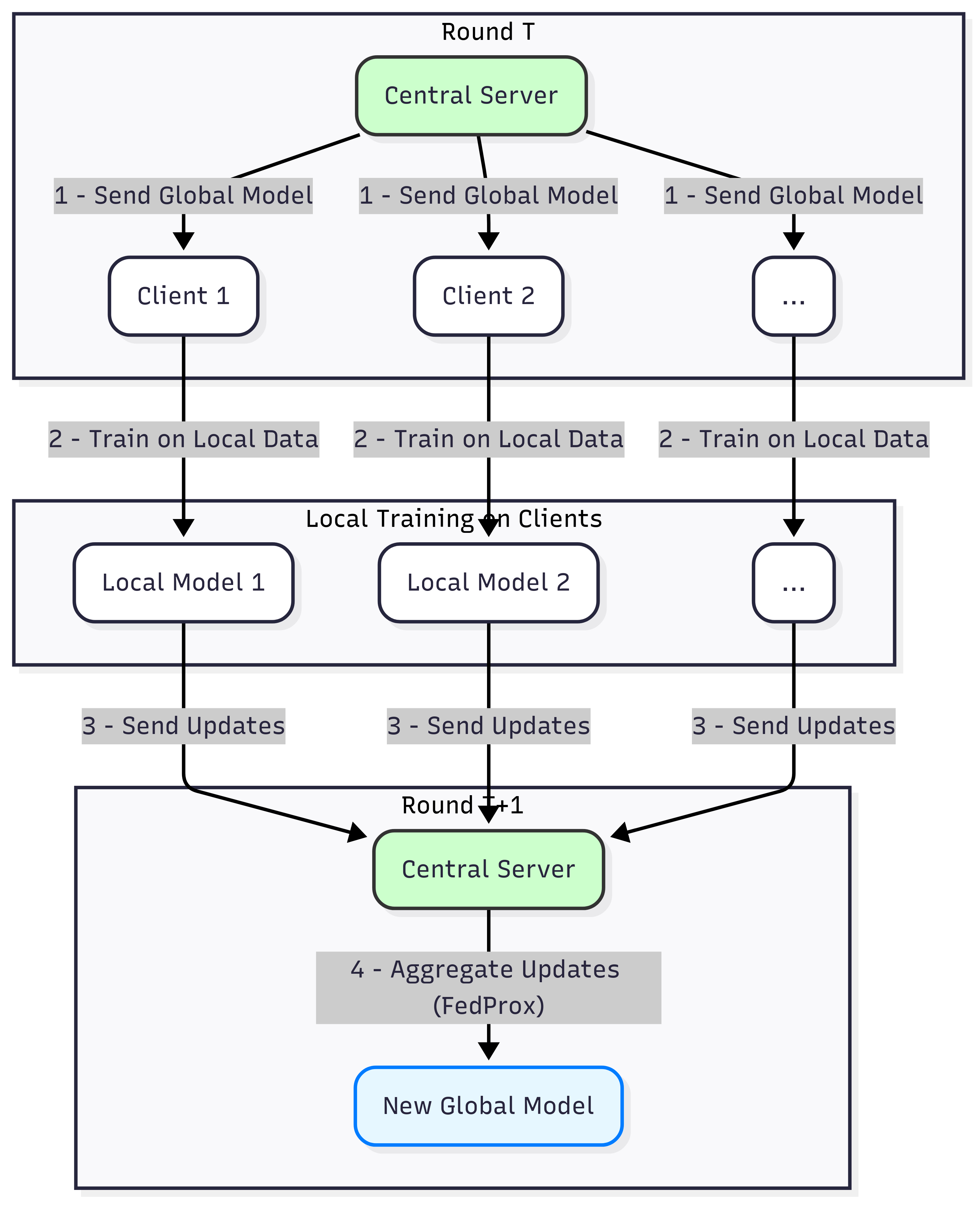}
    \caption{Federated Learning (FedProx) Workflow}
    \label{fig:federated}
\end{figure}

To reconcile the benefits of large-scale data analysis with these stringent privacy requirements, novel computational paradigms are necessary. Among these, Federated Learning (FL), a machine learning paradigm originally developed by researchers at Google in 2026~\cite{Hudaib2025-py}, has emerged as a compelling solution. The methodology was conceived to train models on decentralized data, such as that on mobile devices, without data ever leaving the user's device. Meanwhile, the approach enables collaborative model training across multiple clients (e.g., schools), in this context, educational institutions, without exchanging or centralizing raw data. Instead, a global model is trained iteratively: a server sends the model to each client, each client trains the model on its local private data, and only the resulting model updates (anonymous numerical parameters) are sent back for aggregation. Such a process allows for the creation of a robust global model that learns from the collective data of all institutions. At the same time, sensitive student records remain securely within their original boundaries. The iterative workflow is illustrated in Figure \ref{fig:federated}. A central server orchestrates this process, which allows for the aggregation of model learnings while sensitive information remains securely on each client's local infrastructure.


\begin{enumerate}
    \item \textbf{Initialization:} The server initializes a global model and sends its parameters (weights) to a selection of clients.
    \item \textbf{Local Training:} Each selected client trains the received model on local data for a few iterations.
    \item \textbf{Update Communication:} The clients send only their updated model parameters, not the data, back to the server.
    \item \textbf{Secure Aggregation:} The server aggregates the updates from all clients (e.g., by taking a weighted average of the parameters) to produce an improved global model.
    \item \textbf{Iteration:} Steps 2-4 are repeated for several rounds until the global model converges.
\end{enumerate}

The most fundamental aggregation algorithm in FL is Federated Averaging (FedAvg)~\cite{Khowaja2023}. After clients have trained their local models, the server updates the global model parameters, $w_{t+1}$, for the next round $t+1$ as follows:

\begin{equation}
\label{eq:fedavg_aggregation}
w_{t+1} \leftarrow \sum_{k=1}^{K} \frac{n_k}{N} w_{t+1}^k
\end{equation}

Where $K$ is the total number of clients, $n_k$ is the number of data samples on client $k$, $N$ is the total number of samples across all clients, and $w_{t+1}^k$ are the model parameters received from client $k$ at the end of the local training round.

The iterative and privacy-preserving nature of the FL approach employed in our experiments is outlined in Algorithm \ref{alg:federated}. The algorithm details the cyclical interaction between the central server and the participating clients, including the local optimization with the FedProx objective~\cite{10.1145/3286490.3286559}.

\begin{algorithm}[h!]
\caption{Federated Learning Workflow with FedProx}
\label{alg:federated}
\begin{lstlisting}[language=Python]
# Input: Num of rounds T, set of all clients K, FedProx param mu
# Output: A trained global model with final weights w_T

def Server_Execution(T, K, mu):
    # 1. Initialize global model at the server
    w_global = initialize_global_model()

    for t in range(T):
        # 2. Select a subset of clients for the current round
        S_t = select_clients(K)
        
        # 3. Broadcast the global model to selected clients
        #    and collect their trained updates in parallel
        local_updates = []
        for client_k in S_t:
            update = Client_Execution(client_k, w_global, mu)
            local_updates.append(update)

        # 4. Aggregate local updates to improve the global model
        # Aggregation formula: w_global = sum(n_k/N * w_k)
        w_global = aggregate_updates(local_updates)

    return w_global


def Client_Execution(client_k, w_global, mu):
    # a. Client receives the current global model
    w_local = w_global

    # b. Client trains the model on its local private data,
    #    minimizing the FedProx objective function:
    #    loss = F_k(w) + (mu/2) * ||w - w_global||^2
    w_local_updated = train_on_local_data(w_local, client_k.data)

    # c. Client sends the updated model back to the server
    return w_local_updated
\end{lstlisting}
\end{algorithm}

\subsection{Addressing Statistical Heterogeneity with FedProx}

A primary challenge in real-world FL scenarios is statistical heterogeneity, where the data distributions across clients are not identically and independently distributed (Non-IID)~\cite{9155494}. In our educational context, this means that the student data at one school may differ greatly from that at another. Such heterogeneity can cause the local models of different clients to diverge significantly during training, leading to instability and poor convergence of the global model when using standard FedAvg~\cite{10322899}.

The FedProx algorithm was specifically designed to mitigate this issue~\cite{electronics12204364}. It modifies the local optimization problem on each client by adding a proximal term to its local loss function. This term penalizes the local model's parameters for drifting too far from the global model's parameters received at the beginning of the round. The modified local objective function for each client $k$ becomes:

\begin{equation}
\label{eq:fedprox_objective}
\min_{w} h_k(w) = F_k(w) + \frac{\mu}{2} ||w - w^t||^2
\end{equation}

In this equation, $F_k(w)$ is the original local loss function for client $k$ on its local data. The second term is the proximal term, where $w^t$ represents the parameters of the global model from the previous round (round $t$), and $w$ represents the parameters of the trained local model. The hyperparameter $\mu \geq 0$ controls the degree of penalty; a larger $\mu$ forces the local models to stay closer to the global model, thereby limiting the impact of local data heterogeneity and promoting a more stable and robust convergence. Our study employs FedProx to effectively handle the expected data heterogeneity across different schools in the SAEB dataset.

\section{Methodology}
\label{sec:methodology}

This section details the comprehensive methodology employed to evaluate the feasibility and performance of FL for student performance prediction on the SAEB dataset. The process is organized into four main stages: dataset description, data preprocessing and feature engineering, experimental setup for centralized and federated models, and the evaluation metrics used for comparison. To provide a consolidated overview of the experimental design, the key characteristics of the dataset, features, and modeling parameters are summarized in Table \ref{tab:dataset_summary}.


\begin{table}[tbp]
\centering
\caption{Summary of Dataset and Experimental Setup Characteristics.}
\label{tab:dataset_summary}
\footnotesize
\begin{tabularx}{\textwidth}{@{}lX@{}}
\toprule
\rowcolor{lightgray} 
\textbf{Parameter} & \textbf{Value / Description} \\ 
\midrule
\multicolumn{2}{@{}l}{\textit{Dataset Characteristics}} \\
Data Source & Brazilian Basic Education Assessment System (SAEB) Microdata \\
Total Student Records & 2,087,904 \\
Target Variable & Binary classification of Mathematics Proficiency (\texttt{PROFICIENCIA\_MT}) based on the median score. \\ 
\addlinespace
\multicolumn{2}{@{}l}{\textit{Feature Details}} \\
Predictor Features & 11 selected socioeconomic and demographic variables (e.g., parental education, family income, gender). \\
Number of Features (Post Encoding) & 54 (after One-Hot Encoding). \\ 
\addlinespace
\multicolumn{2}{@{}l}{\textit{Experimental Setup}} \\
Models Compared & 1. Centralized: XGBoost \\ & 2. Federated: DNN with FedProx \\
Number of FL Clients & 50 schools (randomly sampled). \\
FL Client Filter & Minimum of 20 student records per school. \\ 
\bottomrule
\end{tabularx}
\end{table}

\subsection{Dataset Description}

The dataset used in this study comprises microdata from the Brazilian Basic Education Assessment System (SAEB), a nationwide standardized evaluation managed and publicly provided by the National Institute for Educational Studies and Research Anísio Teixeira (INEP) \cite{INEP2025SAEB}. The SAEB dataset is a rich resource, containing student performance metrics in key subjects and a detailed socioeconomic questionnaire. For our analysis, we utilized the data corresponding to 9th-grade students from the most recent available year. The raw dataset is extensive, and after initial loading and filtering, our working dataset for preprocessing consisted of over two million student records, each associated with a specific school identifier (\texttt{ID\_ESCOLA}), which is crucial for the FL analysis.

\subsection{Data Preprocessing and Feature Engineering}

\begin{figure}[h!]
    \centering
    \includegraphics[width=0.3\textwidth]{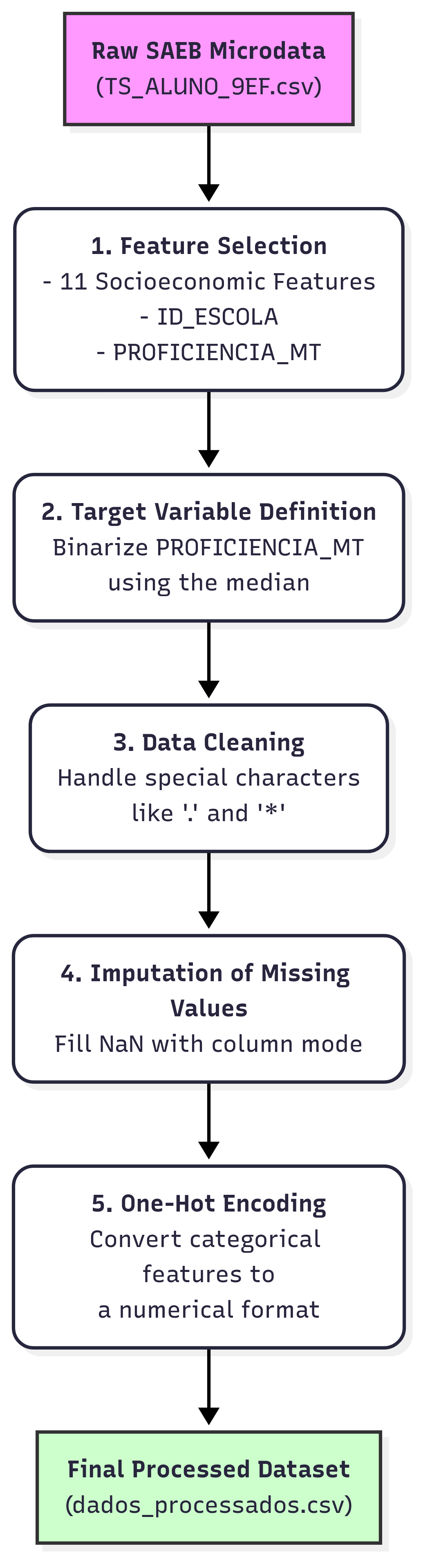}
    \caption{The data preprocessing pipeline, from raw data selection to the final model-ready dataset}
    \label{fig:preprocessing_pipeline}
\end{figure}

A rigorous preprocessing pipeline was designed and implemented to transform the raw SAEB data into a clean, model-ready format. The process, illustrated in Figure \ref{fig:preprocessing_pipeline}, involved four critical steps.

\paragraph{1. Feature Selection} Instead of using all 153 available variables, a subset of 11 socioeconomic and demographic features was selected based on our literature review and exploratory data analysis findings. These features included parental education level, family income, gender, race, and access to resources like internet and computers (e.g., \texttt{TX\_RESP\_Q02}, \texttt{TX\_RESP\_Q06}). The school identifier, \texttt{ID\_ESCOLA}, and the target performance metric, \texttt{PROFICIENCIA\_MT} (mathematics proficiency), were also retained.

\paragraph{2. Target Variable Definition} To frame the task as a binary classification problem, the continuous mathematics proficiency score (\texttt{PROFICIENCIA\_MT}) was transformed into a binary target variable named \texttt{ALVO\_CLASSIFICACAO}. The median of the proficiency score distribution was calculated and used as the threshold. Students with a score above the median were assigned to class `1`, while those with a score less than or equal to the median were assigned to class `0`. This approach naturally created a perfectly balanced dataset with a 50/50 class distribution, obviating the need for data resampling techniques like SMOTE~\cite{11122437}.

\paragraph{3. Data Cleaning and Imputation} The pipeline began by dropping any rows where the target variable (\texttt{PROFICIENCIA\_MT}) was missing. Subsequently, non-standard missing value indicators within the selected features (such as `.` or `*`) were replaced with `NaN` (Not a Number) to ensure consistency. An imputation strategy was applied to handle these missing values. For each categorical feature, the `NaN` values were filled with the column's \textbf{mode} (i.e., the most frequently occurring category), a standard and effective technique for this data type.

\paragraph{4. One-Hot Encoding} The final preprocessing step was to convert the cleaned, categorical features into a numerical format suitable for ML algorithms. One-hot encoding was applied to each of the 11 selected features. Moreover, the process transforms a column with $N$ categories into $N$ new binary columns, reducing the risk of the model incorrectly interpreting ordinal relationships between categories. As a result, the dataset comprised the \texttt{ID\_ESCOLA}, the binary target variable, and 54 one-hot encoded feature columns.

\subsection{Experimental Setup and Models}

The experiment was designed to directly compare the performance of a traditional centralized model against our proposed FL model on the same prediction task.

\subsubsection{Centralized Model (Benchmark)}

The centralized benchmark was established using an \textbf{XGBoost} classifier, renowned for its high performance on tabular data. The preprocessed dataset, containing 2,087,904 student records, was split into a training set (80\%) and a testing set (20\%). A stratified split was used to preserve the 50/50 class distribution in both sets. The XGBoost model was trained on the training data, and its final performance was evaluated on the unseen testing data.

\subsubsection{Federated Learning (Benchmark)}

The FL environment was simulated using the \textbf{Flower} framework~\cite{10.1145/3637528.3671447}. The preprocessed dataset was partitioned by the \texttt{ID\_ESCOLA} column, treating each unique school as an independent client with private data.

\begin{itemize}
    \item \textbf{Client Selection:} To ensure each client had sufficient data for meaningful local training, a filter was applied to include only schools with a minimum of \textbf{20 student records}. A random sample of \textbf{50 schools} was used for the analysis from this pool of valid schools.
    \item \textbf{Model:} A \textbf{Deep Neural Network (DNN)} was implemented using PyTorch. The architecture consisted of an input layer corresponding to the \textbf{54 features}, two hidden layers (64 and 32 neurons, respectively) using the ReLU activation function, and a final output layer with a single neuron and a Sigmoid activation function for binary classification.
    \item \textbf{Federated Algorithm:} The \textbf{FedProx} algorithm was chosen as the aggregation strategy to effectively handle the statistical heterogeneity (Non-IID) of the data across the different schools. The analysis was run for \textbf{10 rounds}, with each participating client performing \textbf{5 local training epochs} per round. The proximal term hyperparameter, $\mu$, was set to $0.1$. A global test set, separated before the analysis, was used to evaluate the global model's performance at the end of each round.
\end{itemize}

\subsection{Evaluation Metrics}

A suite of standard classification metrics was employed to provide a comprehensive and robust assessment of the models. These metrics are derived from the four outcomes of a binary confusion matrix: True Positives (TP), the number of positive instances correctly classified; True Negatives (TN), the number of negative instances correctly classified; False Positives (FP), the number of negative instances incorrectly classified as positive; and False Negatives (FN), the number of positive instances incorrectly classified as negative.

\paragraph{Accuracy} This metric represents the overall proportion of correct predictions among the total number of instances evaluated. It provides a general measure of the model's performance and is calculated as:
\begin{equation}
\label{eq:accuracy}
\text{Accuracy} = \frac{TP + TN}{TP + TN + FP + FN}
\end{equation}
\paragraph{Precision, Recall, and F1-Score} To gain deeper insights beyond overall accuracy, we also evaluated Precision, Recall, and the F1-Score. 
\begin{itemize}
    \item \textbf{Precision} measures the exactness of the model, quantifying the proportion of positive predictions that were actually correct. It is critical in scenarios where False Positives are costly.
   \begin{equation}
\label{eq:precision}
\text{Precision} = \frac{TP}{TP + FP}
\end{equation}
    \item \textbf{Recall} (or Sensitivity) measures the completeness of the model, quantifying the proportion of actual positive instances that were correctly identified. It is crucial that failing to detect a positive case (a False Negative) is a significant issue.
  \begin{equation}
\label{eq:recall}
\text{Recall} = \frac{TP}{TP + FN}
\end{equation}
    \item \textbf{F1-Score} is the harmonic mean of Precision and Recall, providing a single, balanced score that is particularly useful when there is an uneven class distribution or when there is a need to balance the trade-off between Precision and Recall.
\begin{equation}
\label{eq:f1score}
\text{F1-Score} = 2 \times \frac{\text{Precision} \times \text{Recall}}{\text{Precision} + \text{Recall}}
\end{equation}
\end{itemize}

\paragraph{ROC Curve and AUC Score} To evaluate the model's discriminative ability across all classification thresholds, the Receiver Operating Characteristic (ROC) curve was generated. The ROC curve plots the True Positive Rate (Recall) against the False Positive Rate. The Area Under the Curve (AUC) provides a single scalar value summarizing this performance, where an AUC of 1.0 represents a perfect classifier and an AUC of 0.5 represents a model with no discriminative power beyond random chance.

These metrics were applied to the centralized and federated models to directly and comprehensively compare their predictive capabilities.

\section{Results}
\label{sec:results}

This section presents the empirical outcomes of the comparative analysis between the centralized and FL models. The performance of each approach is detailed, focusing on predictive accuracy and model behavior.

\subsection{Centralized Model Performance}

The centralized XGBoost model was trained on the entire preprocessed dataset to establish a performance benchmark. This model represents the theoretical maximum performance achievable when data privacy constraints are disregarded. The final model achieved a predictive accuracy of \textbf{63.96\%} on the unseen test set.

\begin{figure}[h!]
    \centering
    \includegraphics[width=0.6\textwidth]{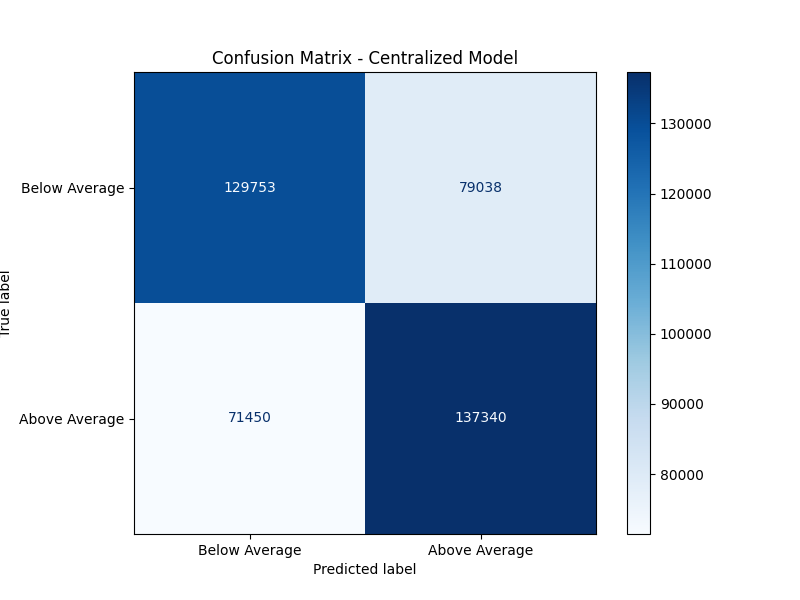}
    \caption{Confusion matrix for the centralized XGBoost model on the test set}
    \label{fig:confusion_centralized}
\end{figure}

Figure \ref{fig:confusion_centralized} presents the confusion matrix for the model's performance. The main diagonal shows that the model correctly classified \textbf{129,753} instances as "Below Average" (True Negatives) and \textbf{137,340} as "Above Average" (True Positives). The off-diagonal values represent the misclassifications, with \textbf{79,038} false positives and \textbf{71,450} false negatives. The relatively balanced distribution of these errors demonstrates that the model does not exhibit significant bias and performs comparably for both classes. Consequently, the similar precision and recall scores in the classification report support this conclusion.

\begin{figure}[h!]
    \centering
    \includegraphics[width=0.9\textwidth]{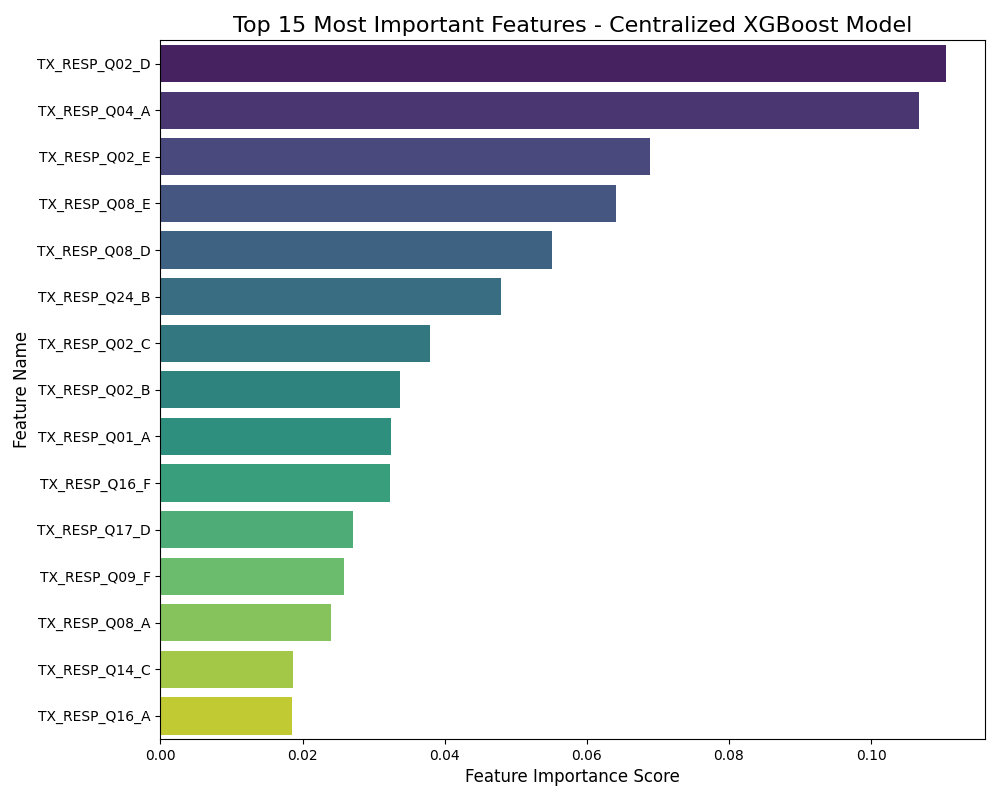}
    \caption{Top 15 most important features identified by the centralized XGBoost model}
    \label{fig:feature_importance_centralized}
\end{figure}

To understand which factors drove these predictions, Figure \ref{fig:feature_importance_centralized} illustrates the top 15 most important features determined by the XGBoost model. The analysis reveals that socioeconomic indicators are the primary drivers of the model's predictions. Notably, specific categories related to family income (e.g., \texttt{TX\_RESP\_Q06\_G}, \texttt{TX\_RESP\_Q06\_F}) and parental education level (e.g., \texttt{TX\_RESP\_Q04\_H}, \texttt{TX\_RESP\_Q05a\_H}) rank among the most influential features. Multiple distinct categories from these questions (representing different income or education brackets) underscore the model's ability to learn nuanced patterns directly from the socioeconomic data provided in the SAEB questionnaire. As a result, confirms that variables related to the students' household environment are key predictors of academic performance in this centralized context. 

To understand which factors drove these predictions, Table \ref{tab:feature_interpretation} decodes the most influential features identified by the XGBoost model. The results clearly show that factors related to family capital, economic (income), and cultural (parental education, presence of books) are the most dominant predictors. Features such as belonging to the highest income bracket or having parents with post-graduate education are strongly associated with higher performance. Conversely, the absence of books at home significantly predicts lower performance. This analysis confirms that the model successfully learned to identify nuanced socioeconomic patterns within the SAEB data to inform its predictions. 

\begin{table}[h!] 
\centering
\caption{Interpretation of the Top 5 Most Important Features Identified by the Centralized XGBoost Model}
\label{tab:feature_interpretation}
\footnotesize 
\begin{tabularx}{\textwidth}{@{}clXX@{}}
\toprule
\rowcolor{lightgray} 
\textbf{Rank} & \textbf{Feature Name} & \textbf{Description (from Data Dictionary)} & \textbf{Hypothesized Relationship with Performance} \\ 
\midrule
1 & \texttt{TX\_RESP\_Q06\_G} & \textbf{Family Income:} More than R\$ 13,200.00. & Higher income is strongly correlated with access to better educational resources (private tutoring, technology, quality housing), positively impacting student performance. \\ 
\addlinespace
2 & \texttt{TX\_RESP\_Q09\_A} & \textbf{Books at Home:} No. & The absence of non-academic books indicates a home environment with fewer stimuli for reading and intellectual curiosity, which can negatively affect academic development. \\ 
\addlinespace
3 & \texttt{TX\_RESP\_Q04\_H} & \textbf{Father's Education:} Post-Graduate degree. & Parents with higher education levels tend to value and provide more support for their children's academic activities, creating a favorable environment for success. \\ 
\addlinespace
4 & \texttt{TX\_RESP\_Q06\_F} & \textbf{Family Income:} From R\$ 7,920.01 to R\$ 13,200.00. & Similar to the highest bracket, this upper-middle income range is a strong positive predictor, indicating significant resource access. \\ 
\addlinespace
5 & \texttt{TX\_RESP\_Q05a\_H} & \textbf{Mother's Education:} Post-Graduate degree. & Corroborates the finding for the father's education, highlighting the immense impact of parental academic background on student achievement. \\ 
\bottomrule
\end{tabularx}
\end{table}


\subsection{Federated Model Performance}

\begin{figure}[h!]
    \centering
    \includegraphics[width=\textwidth]{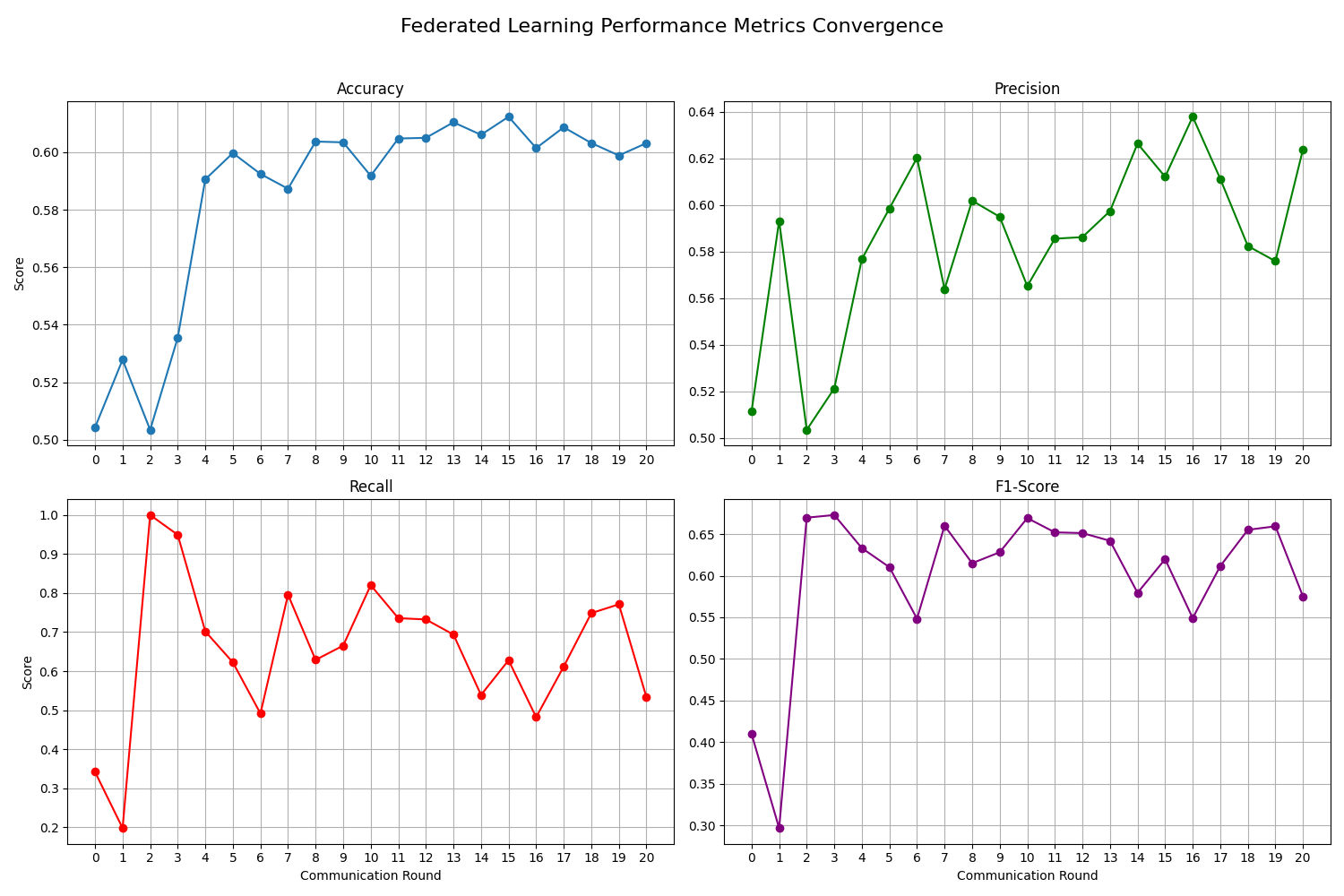} 
    \caption{Global performance metrics of the federated DNN model over 20 communication rounds, showing the convergence of Accuracy, Precision, Recall, and F1-Score.}
    \label{fig:convergence_federated}
\end{figure}

Utilizing a Deep Neural Network with the FedProx algorithm, the federated model was trained over 20 communication rounds to evaluate a privacy-preserving approach. Figure \ref{fig:convergence_federated} depicts the model's learning progress, which plots the convergence of four key performance metrics. The graphs illustrate a successful, albeit volatile, training process. The model's global accuracy improves from an initial state near random chance to a peak performance of \textbf{61.23\%} in round 15, demonstrating that the federated training effectively learned from the distributed and heterogeneous data.

A detailed analysis of the metrics history reveals the nuances of federated training. The model achieved a peak \textbf{F1-Score of 67.29\%} early in the training (round 3), indicating an initial strong balance between Precision and Recall. The \textbf{Precision} peaked later at \textbf{63.79\%} (round 16), while the \textbf{Recall} for the "Above Average" class reached a stable high point of \textbf{94.92\%} (round 3), showing a strong capability to identify true positive cases.

\begin{figure}[h!]
    \centering
    \includegraphics[width=0.7\textwidth]{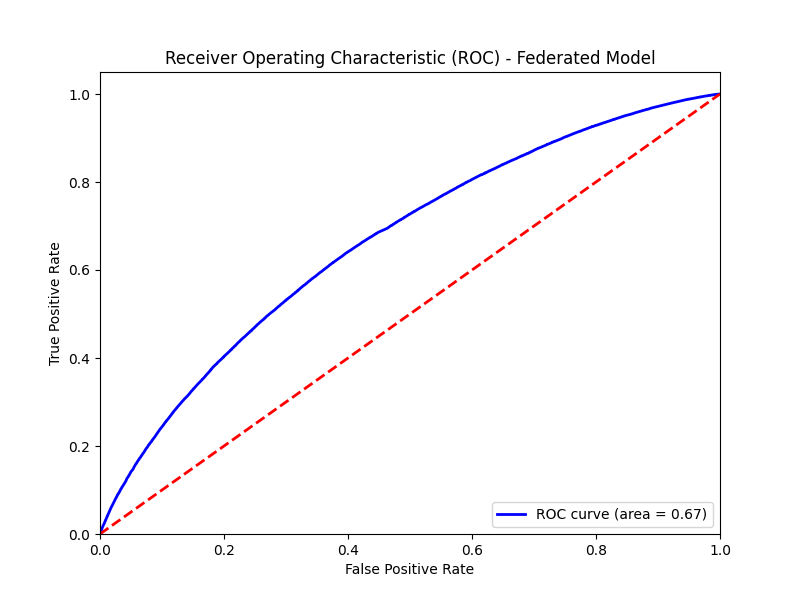}
    \caption{Receiver Operating Characteristic (ROC) curve for the final federated DNN model, evaluated on the global test set after 20 rounds.}
    \label{fig:roc_federated}
\end{figure}

The final discriminative power of the converged federated model is illustrated by its Receiver Operating Characteristic (ROC) curve, shown in Figure \ref{fig:roc_federated}. The Area Under the Curve (AUC) score of \textbf{67.96\%}, calculated on the final model, confirms its meaningful predictive capability in distinguishing between the two student performance classes, reinforcing the viability of the federated approach.

\section{Discussion}
\label{sec:discussion}

The results presented provide the basis for a nuanced discussion on the practical application of FL in a real-world educational context. This section interprets the performance-privacy trade-off, addresses the inherent challenges of working with the SAEB dataset, and considers the implications for educational policy.

\subsection{The Performance-Privacy Trade-off}

The central finding of this study is the quantifiable trade-off between model performance and data privacy. The centralized XGBoost model achieved an accuracy of \textbf{63.96\%}, establishing a robust benchmark for maximum theoretical performance. The privacy-preserving FL model (DNN with FedProx) successfully converged over 20 communication rounds to a peak accuracy of \textbf{61.23\%}. This result is highly significant, as it narrows the performance gap between the two paradigms to a mere \textbf{2.73 percentage points}.

This minimal gap can be interpreted as the "cost of privacy", a marginal and acceptable decrease in predictive power required to ensure that no sensitive student data is ever centralized. We argue that this trade-off is highly favorable for practical application. In a legal landscape governed by the LGPD, the alternative to a privacy-preserving collaborative model is often no model at all. Therefore, achieving an accuracy of over \textbf{61\%} in a fully privacy-compliant manner is a substantial advancement. It transforms a previously intractable problem into a solvable one, enabling data-driven insights where they were previously impossible to obtain legally and ethically.

\subsection{Answering the Research Questions}

The preceding sections have detailed the methodology, presented the empirical results, and discussed their implications. Based on this comprehensive analysis, we now directly answer the research questions that motivated this study.

\begin{enumerate}    
\item \textbf{Answering RQ1:} To what extent can socioeconomic features within the SAEB dataset predict student
academic performance using a centralized, state-of-the-art ML model.

Our first research question (RQ1) asked to what extent socioeconomic features within the SAEB dataset can predict student academic performance using a centralized model. The results demonstrate a significant and measurable predictive power. Our state-of-the-art centralized model (XGBoost), when trained on the complete aggregated dataset, achieved an accuracy of \textbf{63.96\%}. As shown in Figure \ref{fig:confusion_centralized}, the model's balanced performance across both classes indicates that the selected socioeconomic features contain a clear and useful signal. This outcome confirms that, under a traditional data aggregation paradigm, it is possible to build a model that can identify at-risk students with an efficacy substantially better than random chance, providing a robust performance benchmark for our study.

\item \textbf{Answering RQ2:} What is the predictive performance of a Deep Neural Network trained via a Federated
Learning approach (FedProx) on the same prediction task, without centralizing sensitive
student data.

The second research question (RQ2) investigated the predictive performance of a privacy-preserving federated model. Our analysis demonstrated successful convergence and robust performance by employing a Deep Neural Network (DNN) with the FedProx algorithm. The model achieved a final accuracy of \textbf{60.32\%} after 20 communication rounds, with a peak performance of \textbf{61.23\%} during training (Figure \ref{fig:convergence_federated}). Furthermore, the model's ability to discriminate between classes was confirmed by an Area Under the Curve (AUC) score of \textbf{0.6796}, as depicted in Figure \ref{fig:roc_federated}. This result is highly significant, as it establishes that training an effective and robust predictive model on the same task is feasible without centralizing sensitive student data, thereby respecting the core principles of data privacy. The successful convergence of the federated model provides a strong affirmative answer to this research question.

\item \textbf{Answering RQ3:} How significant is the performance trade-off between the privacy-preserving federated model and the centralized benchmark, and is such a trade-off acceptable for practical
application in the Brazilian educational context

Our third research question (RQ3) addressed the significance of the performance trade-off and its acceptability in the Brazilian educational context. The performance gap between the centralized benchmark of \textbf{63.96\%} and the federated model's peak accuracy of \textbf{61.23\%} is approximately \textbf{2.73 percentage points}. The trade-off is acceptable and highly favorable for practical application. In a legal landscape governed by the LGPD, the alternative to a privacy-preserving collaborative model is often no model at all. The ability to achieve over \textbf{61\%} accuracy in a privacy-compliant manner is a transformative step, enabling data-driven insights that were previously unachievable.

\end{enumerate}

\subsection{The Challenge of Predicting Performance with SAEB Data}

A crucial aspect of this discussion is understanding why the model's performance, with accuracy converging to approximately \textbf{61-62\%} while other metrics like F1-Score and Recall showed greater volatility, did not achieve higher levels. Hence, it is not a limitation of the models, but a reflection of the inherent complexity of the educational data itself. 

\begin{itemize}
    \item \textbf{High Human and Social Variability:} Student academic performance is a deeply multifactorial phenomenon. It is influenced by many variables not captured in the SAEB questionnaire, such as individual student motivation, teacher quality, specific pedagogical approaches, and personal life events. These unobserved variables introduce a significant amount of natural variance, or "noise," which creates a ceiling on the theoretical maximum predictability.
    
    \item \textbf{Socioeconomic Proxies vs. Deterministic Causes:} The features used in our model (e.g., parental education, income) are powerful statistical \textbf{proxies}, not deterministic causes. They indicate a higher or lower \textit{probability} of a certain outcome across a population, but they do not determine the outcome for any single individual. A predictive model trained on such data will inevitably make errors when individual circumstances defy the general trend.
    
    \item \textbf{Cross-Sectional Data:} The SAEB dataset is a cross-sectional "snapshot" of students at a single point in time. It does not provide a longitudinal view of their academic progress or engagement over several years. Models trained on such data can capture correlations but struggle to model the dynamic, evolving nature of a student's educational journey.
\end{itemize}

Given these factors, achieving an accuracy in the low-to-mid 60s is a strong and realistic result. It confirms that the socioeconomic data in SAEB contains a significant predictive signal, despite the inherent complexities and human nature of education that no dataset can fully capture.

\subsection{Implications for Educational Policy}

While not intended for high-stakes decisions about individual students, the predictive models developed in this study are exceptionally valuable as tools for \textbf{policy and resource allocation}. An educational system could use the federated model to generate school risk profiles without accessing student-level data. Schools with more at-risk students could be targeted for systemic interventions, such as increased funding or specialized teacher training programs. The model thus functions not as a tool for judging individuals, but as a "macroscope" for identifying and addressing systemic inequalities in a data-informed and privacy-preserving manner.

\section{Conclusion}
\label{sec:conclusion}

The study was motivated by the critical challenge of leveraging large-scale educational data to generate predictive insights while adhering to stringent data privacy regulations such as Brazil's LGPD. The research sought to answer whether a privacy-preserving Federated Learning (FL) approach could achieve a predictive performance comparable to a traditional, centralized model on a real-world, large-scale national dataset. We provide a clear and affirmative answer through a comprehensive methodological pipeline involving data preprocessing, modeling, and simulation on over two million student records from the SAEB.

Our findings demonstrate that centralized and federated paradigms can successfully model the relationship between students' socioeconomic backgrounds and academic performance. The centralized XGBoost model established a strong performance benchmark with an accuracy of \textbf{63.96\%}. More significantly, using a DNN with the FedProx algorithm, our optimized FL model successfully converged over 20 communication rounds to a peak accuracy of \textbf{61.23\%}. This result is paramount, as it quantifies the performance-privacy trade-off to a mere \textbf{2.73 percentage points}. Such a minimal performance gap strongly supports the conclusion that FL is a viable and highly effective approach for this task.

The primary contribution of this work is the empirical demonstration that collaborative, privacy-compliant machine learning is feasible in the Brazilian educational context. We provide a practical blueprint and a robust performance benchmark to guide future research and institutional initiatives. Our results show that building a powerful "macroscope" to identify systemic risks and inform policy without compromising the fundamental right to student data privacy is possible.

\section{Future Work}
\label{sec:future_work}

While this study establishes a strong foundation, several promising avenues for future research emerge, directly inspired by recent advancements in the literature.

\begin{itemize}
    \item \textbf{Integration of Explainable AI (XAI):} Our current model provides accurate predictions, but its internal decision-making process remains a "black box". Future work should focus on integrating XAI techniques, such as Shapley Additive exPlanations (SHAP), into the federated framework. As demonstrated by~\textcite{Mastour2025}, XAI can provide granular, instance-level explanations, transforming predictive models into trustworthy and actionable tools for educators to understand the specific factors influencing a student's risk profile.

    \item \textbf{Enrichment with VLE Data and Feature Engineering:} This study was based on cross-sectional socioeconomic data. The literature, particularly the work of~\textcite{Junejo2025} and~\textcite{Marcolino2025}, highlights the immense predictive value of longitudinal data from Virtual Learning Environments (VLEs), such as clickstream and engagement logs. A significant next step would be to design federated models where each client (school) can locally incorporate VLE data, potentially creating more powerful and timely features for early-stage prediction.

    \item \textbf{Exploration of Advanced FL Algorithms and Architectures:} We employed the robust FedProx algorithm. However, the field of FL is rapidly evolving. Future research could explore more advanced aggregation strategies or alternative model architectures. The consistent success of tree-based models in the literature, evidenced by~\textcite{Delena2025} and our own centralized benchmark, suggests that the implementation of a Federated XGBoost framework remains a highly promising, albeit more complex, research direction that could potentially close the performance gap even further.
\end{itemize}



\section*{Acknowledgements}
The author would like to acknowledge the support of the Software Engineering and Automation Research Laboratory (LaPEA), where the research was developed and conducted. The infrastructure and resources provided were essential for the completion of the work.


\printbibliography

\appendix
\section*{Appendix: Condensed Source Code}
\label{appendix:source_code}

This appendix provides condensed versions of the key Python scripts used for the centralized and federated model experiments. The code highlights the main components of each experiment, including model instantiation, training logic, and evaluation, to ensure methodological clarity and reproducibility without presenting the full boilerplate code. All data in the article can be obtained from the SAEB database, publicly available at: \url{https://www.gov.br/inep/pt-br/acesso-a-informacao/dados-abertos/microdados/saeb} (accessed on August 25, 2025).

\subsection*{Centralized Model Training}
\label{appendix:centralized}

\begin{lstlisting}[language=Python, caption={Key sections of the centralized XGBoost model training script (\texttt{treinamento\_centralizado.py}).}, label={lst:centralized_condensed}]
import pandas as pd
import xgboost as xgb
from sklearn.model_selection import train_test_split
from sklearn.metrics import accuracy_score, classification_report

def treinar_modelo_centralizado():
    # --- 1. DATA LOADING AND PREPARATION ---
    df = pd.read_csv('data/dados_processados.csv')
    
    y = df['ALVO_CLASSIFICACAO']
    X = df.drop(columns=['ALVO_CLASSIFICACAO', 'ID_ESCOLA'])
    
    X_train, X_test, y_train, y_test = train_test_split(
        X, y, test_size=0.2, random_state=42, stratify=y
    )

    # --- 2. XGBOOST MODEL TRAINING ---
    model = xgb.XGBClassifier(
        objective='binary:logistic',
        eval_metric='logloss',
        use_label_encoder=False,
        random_state=42
    )
    model.fit(X_train, y_train)

    # --- 3. PERFORMANCE EVALUATION ---
    y_pred = model.predict(X_test)
    accuracy = accuracy_score(y_test, y_pred)
    
    print(f"\n>> Model Accuracy: {accuracy * 100:.2f}% <<")
    print("\nDetailed Classification Report:")
    print(classification_report(y_test, y_pred, target_names=['Below Average', 'Above Average']))

if __name__ == '__main__':
    treinar_modelo_centralizado()
\end{lstlisting}

\subsection*{Federated Model analysis}
\label{appendix:federated}

\begin{lstlisting}[language=Python, caption={Key sections of the federated model analysis script (\texttt{treinamento\_federado.py}).}, label={lst:federated_condensed}]
import flwr as fl
import torch
import torch.nn as nn
import pandas as pd
from sklearn.model_selection import train_test_split
from sklearn.metrics import classification_report, roc_auc_score
from collections import OrderedDict
import numpy as np

# --- ANALYSIS CONFIGURATIONS ---
NUM_ROUNDS = 20
LOCAL_EPOCHS = 10
PROXIMAL_MU = 0.1

# --- DEEP NEURAL NETWORK ARCHITECTURE (DNN) ---
class Net(nn.Module):
    def __init__(self, input_size, hidden_size=64):
        super(Net, self).__init__()
        self.fc1 = nn.Linear(input_size, hidden_size)
        self.relu1 = nn.ReLU()
        self.fc2 = nn.Linear(hidden_size, hidden_size // 2)
        self.relu2 = nn.ReLU()
        self.fc3 = nn.Linear(hidden_size // 2, 1)
        self.sigmoid = nn.Sigmoid()

    def forward(self, x):
        # ... forward pass logic ...
        return x

# --- FEDERATED CLIENT LOGIC ---
class SaebClient(fl.client.NumPyClient):
    # ... __init__, get_parameters, set_parameters methods ...

    def fit(self, parameters, config):
        self.set_parameters(parameters)
        # ... optimizer and criterion logic ...
        self.model.train()
        for _ in range(LOCAL_EPOCHS):
            for features, labels in self.trainloader:
                # ... local training loop ...
        return self.get_parameters(config={}), len(self.trainloader.dataset), {}

# --- SERVER-SIDE EVALUATION FUNCTION ---
def get_evaluate_fn(model, test_loader, total_rounds):
    def evaluate(server_round, parameters, config):
        # ... logic to load weights and evaluate on the global test set ...
        
        accuracy = correct / total if total > 0 else 0
        
        # In the final round, calculate detailed metrics
        if server_round == total_rounds:
            print("\n--- Detailed Evaluation of Final Federated Model ---")
            # ... logic to generate classification report and AUC score ...
        
        return loss, {"accuracy": accuracy}
    return evaluate

# --- SERVER ORCHESTRATION ---
def main():
    # ... data loading and preparation logic ...

    # Define the FedProx strategy, passing the evaluation function
    strategy = fl.server.strategy.FedProx(
        fraction_fit=0.2,
        min_fit_clients=10,
        min_available_clients=clients_to_sample,
        evaluate_fn=get_evaluate_fn(global_model, test_loader, NUM_ROUNDS),
        proximal_mu=PROXIMAL_MU
    )

    # Start the federated analysis
    history = fl.analysis.start_analysis(
        client_fn=client_fn,
        num_clients=clients_to_sample,
        config=fl.server.ServerConfig(num_rounds=NUM_ROUNDS),
        strategy=strategy,
    )
    
    # ... logic to print final accuracy history from the 'history' object ...

if __name__ == "__main__":
    main()
\end{lstlisting}
\end{document}